\documentclass[letterpaper, 10 pt, conference]{ieeeconf}  

\IEEEoverridecommandlockouts 
\usepackage[nolist, nohyperlinks, printonlyused, withpage]{acronym}
\usepackage{booktabs}
\usepackage{cite}
\usepackage{amsmath,amssymb,amsfonts}
\usepackage{algorithmic}
\usepackage[T1]{fontenc}  
\usepackage[utf8]{inputenc}
\usepackage{graphicx}
\usepackage{multirow}
\usepackage[caption=false,lofdepth,lotdepth]{subfig}
\usepackage{textcomp}
\usepackage{xcolor}
\def\BibTeX{{\rm B\kern-.05em{\sc i\kern-.025em b}\kern-.08em
    T\kern-.1667em\lower.7ex\hbox{E}\kern-.125emX}}

\newcommand{\qv}{\boldsymbol{q}}
\newcommand{\ql}{\underline{\qv}}  
\newcommand{\qu}{\overline{\qv}}  
\newcommand{\za}{\mathrm{z}}  
\newcommand{\vecabs}{\|}
\newcommand{\meter}{\text{m}}

\DeclareMathOperator*{\argmin}{arg\,min}

\usepackage{hyperref}  

\begin{document}

\title{\LARGE \bf
Optimizing Modular Robot Composition:\\A Lexicographic Genetic Algorithm Approach
}

\makeatletter
\author{%
Jonathan K\"ulz
and Matthias Althoff%
\thanks{The authors are with the Department of Computer Engineering, Technical University of Munich, 85748 Garching, Germany.
{\tt\small jonathan.kuelz@tum.de, althoff@tum.de}.
}}
\makeatother

\begin{acronym}
    \acro{modrob}[MR]{modular robot}
    \acro{fk}[FK]{forward kinematics}
    \acro{ga}[GA]{genetic algorithm}
    \acro{ik}[IK]{inverse kinematics}
    \acro{ompl}[OMPL]{Open Motion Planning Library }
    \acro{rl}[RL]{deep reinforcement learning}
    \acro{tcp}[TCP]{tool center point}
\end{acronym}
\maketitle

\begin{abstract}
Industrial robots are designed as general-purpose hardware with limited ability to adapt to changing task requirements or environments.
Modular robots, on the other hand, offer flexibility and can be easily customized to suit diverse needs.
The morphology, i.e., the form and structure of a robot, significantly impacts the primary performance metrics acquisition cost, cycle time, and energy efficiency.
However, identifying an optimal module composition for a specific task remains an open problem, presenting a substantial hurdle in developing task-tailored modular robots.
Previous approaches either lack adequate exploration of the design space or the possibility to adapt to complex tasks.
We propose combining a genetic algorithm with a lexicographic evaluation of solution candidates to overcome this problem and navigate search spaces exceeding those in prior work by magnitudes in the number of possible compositions.
We demonstrate that our approach outperforms a state-of-the-art baseline and is able to synthesize modular robots for industrial tasks in cluttered environments.
\end{abstract}

\section{Introduction}
\begin{figure*}
    \centering
    \hfill
        \subfloat[For narrow orientation tolerances at the goal positions, our optimization yields a robot with six degrees of freedom.]{\includegraphics[width=0.48\linewidth,trim={660px 350px 720px 320px},clip]{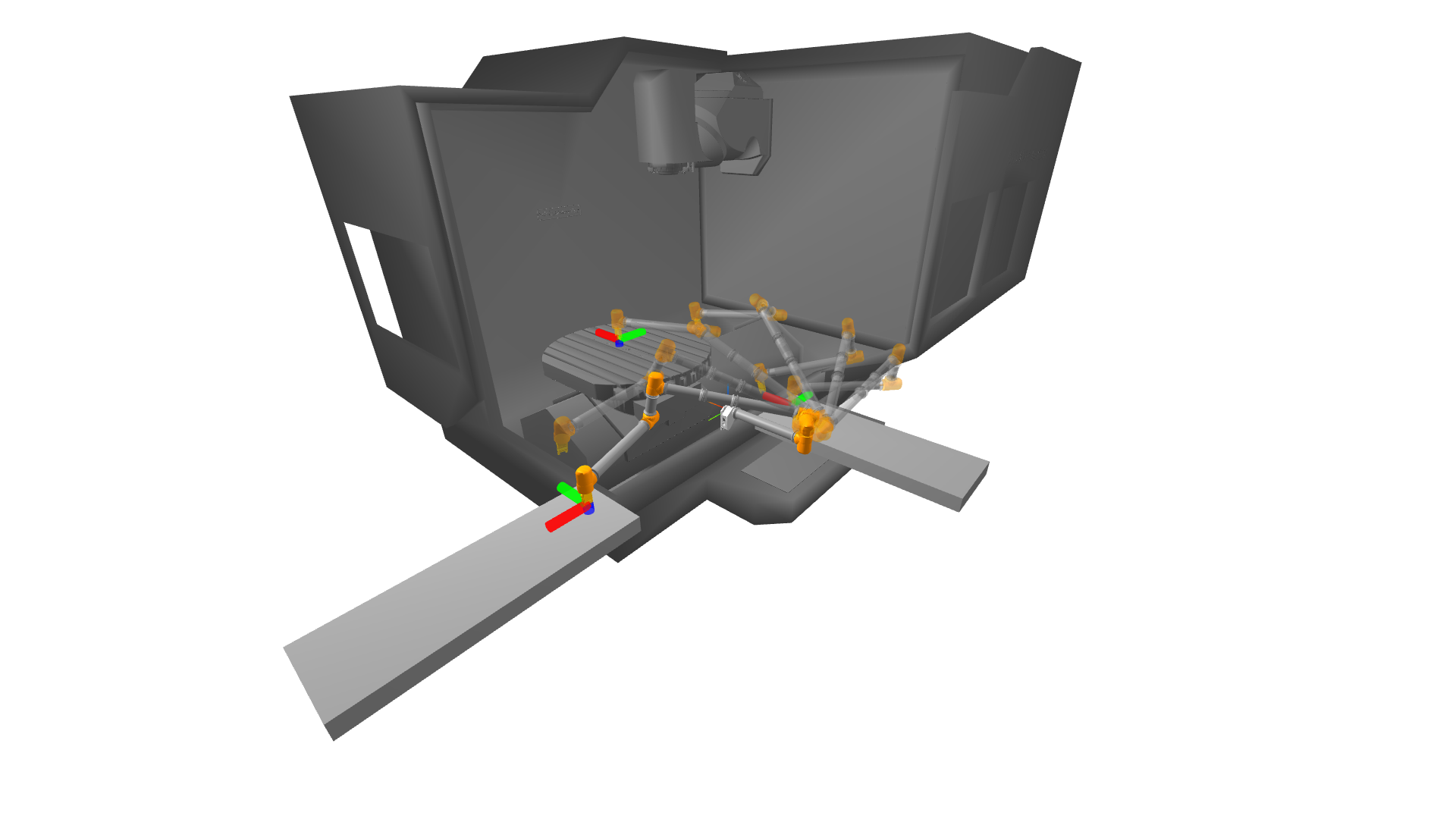}} \label{fig:trajectory_sol_a}
    \hfill
            \subfloat[An \ac{modrob} with just four degrees of freedom solves the task if the orientation tolerances at the desired goal poses are broad.]{\includegraphics[width=0.48\linewidth,trim={650px 350px 730px 320px},clip]{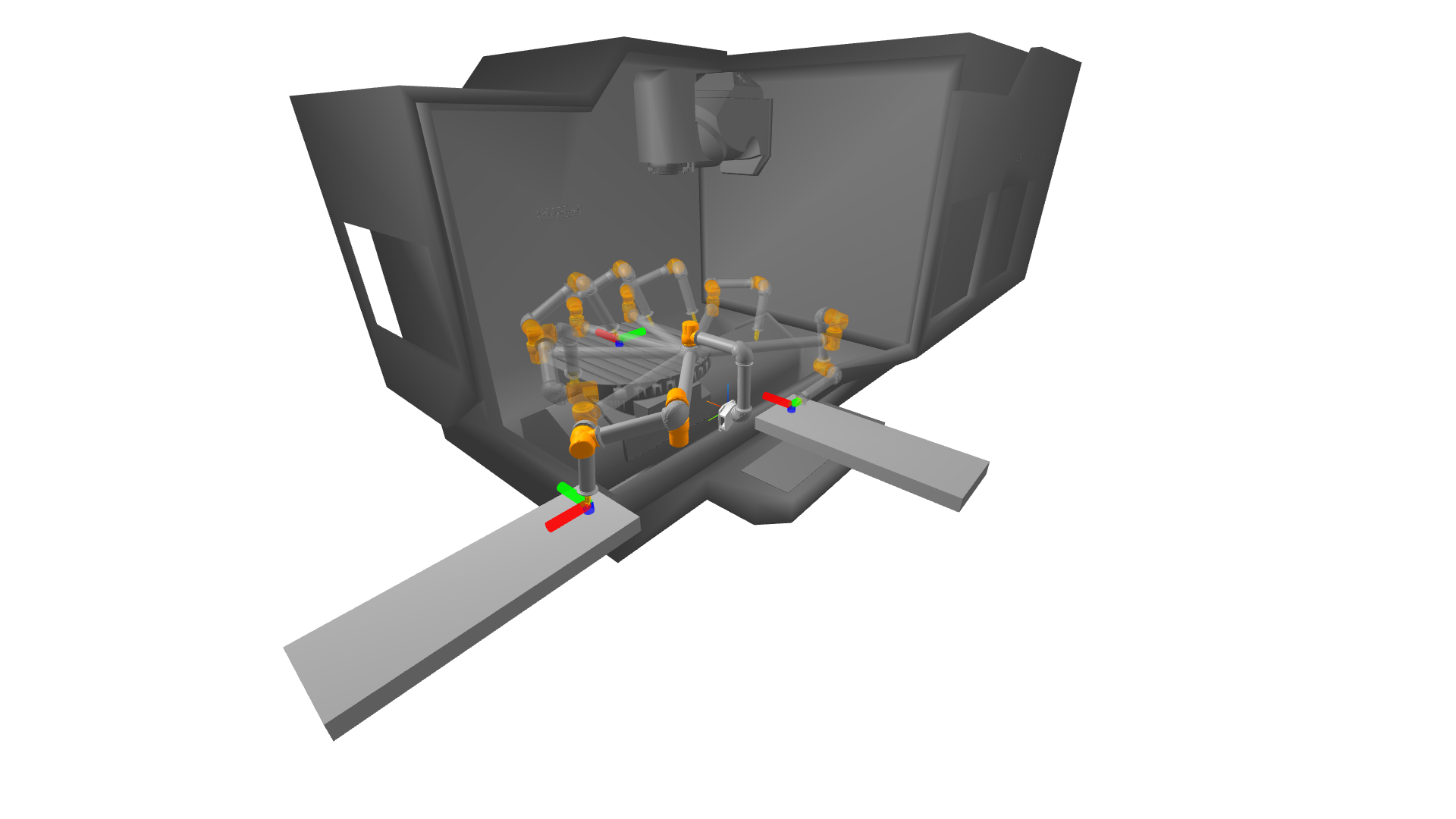}} \label{fig:trajectory_sol_b}
    \caption{
        Adapting task-tailored \ac{modrob} compositions leads to optimized setup and production costs:
        In the example shown above, a small change in the task requirements results in two robots of significantly different complexity.
        While a six-degree-of-freedom robot is necessary to reach all desired goal positions exactly (a), a relaxation of orientation tolerances allows us to design an \ac{modrob} with four degrees of freedom only to solve the task (b).
    }
    \label{fig:trajectory_sol}
\end{figure*}

\Acfp{modrob} are an intuitive solution to an increasing need for mass customization and small-scale manufacturing~\cite{Pedersen2015, Gorecky2016, Ribeiro2018}.
Thanks to their versatility and robustness by design, modular manipulators promise significant technological advances in industrial automation~\cite{Yim2007, Althoff2019}.
As shown in Fig.~\ref{fig:trajectory_sol}, diverse tasks require different compositions of robot modules.
One of the main challenges with \acp{modrob} is determining a robot composition, i.e., selecting and arranging modules that suit the given requirements.
Due to the exponential number of possible permutations of modules, an exhaustive search for the optimal modular composition is usually infeasible.
Meta-heuristics such as deep reinforcement learning or \acfp{ga} have been successfully adapted to facilitate \ac{modrob} design optimization~\cite{Whitman2020, Icer2017}.
However, these approaches require evaluating an immense number of robot designs, rendering their straightforward application to complex tasks impossible and resulting in long run times.

\subsection{Contribution}
We propose a novel \ac{ga}-based approach for \ac{modrob} composition optimization on heterogeneous modules driven by a lexicographic fitness function.
By hierarchically assessing the fitness of solution candidates, we combine the exploratory capabilities of \acp{ga} with the computational efficiency of hierarchical elimination methods while enhancing interpretability compared to a single scalar fitness value.
As a result, in contrast to prior approaches, our \ac{ga} \mbox{(a) considers} computationally expensive metrics, such as trajectory cycle time, during optimization, (b) works in arbitrary environments with any number of workspace targets, and (c) imposes no constraints on the robot structure except that a serial kinematic has to be assembled.
We shot that our approach finds module compositions and trajectories in cluttered and complex environments tailored to task-specific requirements and outperforms a state-of-the-art benchmark regarding robot complexity and cycle time.
All tasks and solutions discussed are available in the CoBRA benchmark~\cite{Mayer2024}, together with several interactive  visualizations\footnote{Visit \href{https://cobra.cps.cit.tum.de/tasks?version=2022&scenario_id__icontains=kuelz_lexicographic_ga}{\texttt{https://cobra.cps.cit.tum.de/tasks}} and search for tasks with the name \texttt{kuelz\_lexicographic\_ga}.}.
The project's website and additional videos can be found at \mbox{\href{https://lexicographic-ga.cps.cit.tum.de}{\texttt{https://lexicographic-ga.cps.cit.tum.de}}}.

\subsection{Related Work}
Modular robots were introduced decades ago, starting in the 1980s~\cite{Pritschow1986, Fukuda1988, Schmitz1988}.
While the automatic generation of kinematic and dynamic models~\cite{Nainer2021, Zhang2022} and their self-programming~\cite{Althoff2019} has incentivized the introduction of new robot hardware~\cite{Yun2020, Romiti2022, Dogra2022a}, the optimization of their modular composition still poses a considerable challenge.

As exhaustive approaches usually fail, one potential solution to finding optimized compositions is limiting the search space, for example, by focusing on morphologies commonly seen in monolithic manipulators~\cite{Liu2020}.
Alternatively, meta-heuristics, such as simulated annealing, can be applied to deal with the large search space~\cite{Paredis1993}.
Mixed integer programming~\cite{Valente2016} or an augmented Lagrangian technique~\cite{Singh2018} were introduced to optimize for the static reachability of \ac{modrob} without considering possible robot motion.
The work in~\cite{Whitman2020} deploys a heuristic search leveraging reinforcement learning to solve tasks with multiple goals and obstacles.
However, due to the large number of evaluations necessary for training, the authors have to simplify the tasks by discretizing the space occupied by obstacles, training the agent on tasks with one goal only, and solving for reachability only instead of performing motion planning.

By applying hierarchical elimination, i.e., evaluating robots on a sequence of increasingly computationally expensive criteria, the authors of~\cite{Icer2016, Icer2016a, Althoff2019} significantly speed up the assessment of the feasibility of a particular \ac{modrob} for a given task.
However, these works fail to navigate the vast search space effectively, as they do not harness the information acquired during hierarchical elimination, nor do they employ information about the robots themselves strategically to guide the search process, resulting in the necessity of imposing structural constraints on the module composition.

\Acfp{ga}~\cite{Goldberg1988} are another frequently used approach to optimizing \ac{modrob} composition.
By mimicking natural evolutionary processes, genetic algorithms can be utilized to maximize their fitness for a given environment~\mbox{\cite{Sims1994, Lipson2000}}.
In a \ac{ga}, a gene sequence composes a chromosome representing a solution candidate.
Starting from an initial population, a new generation of chromosomes is recursively created by applying crossover, mutation, and selection operations to the previous population.
Previous work, such as in~\cite{Han1997}, has explored \ac{ga}-based optimization of robot compositions.
However, this research (a) focuses on a limited subset of robot modules, (b) does not consider different or mobile bases, (c) imposes strict constraints on their structure, i.e., the number of joints and the ordering of joint and link modules, and (d) simplifies performance metrics and omits complex factors like the computation of a workspace trajectory, which is crucial in real-world applications.
These challenges are common in related studies~\cite{Chocron1997, Chocron2008, Whitman2020, Kuelz2023}.
In contrast, the authors of~\cite{Icer2017} propose a two-step approach involving the elimination of infeasible \acp{modrob} through hierarchical elimination.
This approach realizes motion planning on a reduced set of promising \acp{modrob}.
Still, it imposes constraints on the number of robot modules and their degrees of freedom to manage the search space.

\section{Problem Statement}
Throughout this work, we aim to find a composition of heterogeneous modules that can be assembled to a robot to solve a given task.
We assume that a module $m$ has $b$ bodies connected by $b-1$ joints and proximal and distal connectors $c_p$ and $c_d$, respectively, defining interfaces to connect it to other modules.
Our method explicitly permits the inclusion of empty bodies, enabling the creation of more complex module structures.
For instance, this flexibility allows one to combine two linear joints with an empty body to construct a planar joint, which can be used to model a mobile base module.
We consider two special types of modules, bases and end effectors.
For base modules, $c_{p}$ represents the reference frame of the base; for end effector modules, $c_{d}$ defines the \acf{tcp}.
Any connector is attached to a body and has a fixed type.
The corresponding modules can be connected if a distal and a proximal connector have the same type.
In this work, we consider any \ac{modrob} that adheres to the following structure:
\begin{center}
    \textit{Base - $M$ - ... - $M$ - End Effector}
\end{center}
Here, $M$ are arbitrary regular modules, so the structure fits any serial manipulator with exactly one base and end effector module.
An assembled composition of $n_M$ modules is referred to as robot $R = (m_1, \dots, m_{n_M} )$.

A task $T$, as displayed in Fig.~\ref{fig:task_visualizations}, is defined by the tuple $\langle G, \mathcal{T}, \mathcal{O} \rangle$, where \mbox{$G = (g_1, \dots, g_{n_G}) $} is a sequence of goals, $\mathcal{T}$ is a set of tolerances for all goals, and $\mathcal{O}$ is the workspace occupied by obstacles.
Joint limits for a robot with $n_J$ joints are given by lower limits \mbox{$\ql \in \mathbb{R}^{n_J}$} and upper limits \mbox{$\qu \in \mathbb{R}^{n_J}$}.
We define the set of all valid joint configurations as \mbox{$\mathcal{Q} = \left[ \ql, \qu \right]$}.
A goal $g \in G \subset SE3$ is defined as a position $\boldsymbol{p}_g \in \mathbb{R}^3$ and a desired orientation $\boldsymbol{n}_g \in SO3$.
We introduce the operator
\begin{align}  
    rot(\boldsymbol{n_1}, \boldsymbol{n_2}) = \langle \boldsymbol{e}, \theta \rangle, \vecabs \boldsymbol{e} \vecabs_2 = 1, \theta \in [0, \pi]
\end{align}
that returns the rotation from $\boldsymbol{n_1}$ to $\boldsymbol{n_2}$ in axis-angle representation, defined as a unit vector $\boldsymbol{e}$ and an angle $\theta$.
Further, we define the \acf{fk} for robot $R$ with joint angles $\qv$ as
\begin{align}
    [\boldsymbol{p}_\text{TCP}, \boldsymbol{n}_\text{TCP}] &= FK(R, \qv)\, ,
\end{align}
with \ac{tcp} position \mbox{$\boldsymbol{p}_\text{TCP} \in \mathbb{R}^3$} and \ac{tcp} orientation \mbox{$\boldsymbol{n}_\text{TCP} \in SO3$}.
The workspace occupied by the robot is given by $\mathcal{A}(R, \qv)$.
The tolerances $\mathcal{T}$ are composed of a position tolerance $t_p \in \mathbb{R}_+$ and an orientation tolerance, given as the tuple \mbox{$t_o = \langle \boldsymbol{t}, \varphi \rangle, \boldsymbol{t} \in \{ \boldsymbol{x} \in \mathbb{R}_+^3 \mid \vecabs\boldsymbol{x}\vecabs_\infty \leq 1 \}, \varphi \in (0, \pi]$}, where $\boldsymbol{t}$ defines the upper bound on the absolute value of the Euler axis of rotation between $\boldsymbol{n}_\text{TCP}$ and $\boldsymbol{n}_g$~\eqref{eq:goal_reached}.

A goal $g$ can be be reached by robot $R$, if there exists a joint configuration $\qv \in \mathcal{Q}$ such that
\begin{align}
    \vecabs \boldsymbol{p}_\text{TCP}(R, \qv) - \boldsymbol{p}_g\vecabs_2 \leq t_p \land \theta~|\boldsymbol{e}| \leq \varphi~\boldsymbol{t}\, , \label{eq:goal_reached}
\end{align}
where $\langle \boldsymbol{e}, \theta \rangle = rot(\boldsymbol{n}_g, \boldsymbol{n}_\text{TCP})$.
In~\eqref{eq:goal_reached}, $|\boldsymbol{e}|$ denotes the element-wise absolute value and the inequality holds if it is true for all dimensions.
We denote this predicate as $r(R, \qv, g, \mathcal{T})$, which evaluates to \texttt{true} if robot $R$ reaches goal $g$ with the joint configuration $\qv$ and \texttt{false} otherwise.
We also introduce the predicate $\hat{r}$ that determines whether a trajectory $\qv(t)$ reaches all goals in the order specified by $G$ as
\begin{align}
    &\hat{r}(R, \qv(t), G, \mathcal{T}) = \exists \left( t_1, \dots, t_n \right) :  \label{eq:goals_in_order} \\
                                &\qquad t_1 \leq t_2 \leq \dots \leq t_{n} = t_{max} \nonumber \\
                                &\qquad\land \forall i \in \{1, \dots, n\}: r(R, \qv(t_i), g_i, \mathcal{T})\, . \nonumber
\end{align}

A task $T$ is considered to be achieved by robot $R$, if there exists a trajectory $\qv(t)$ that is collision-free~\eqref{eq:traj_collision_free}, for which joint velocity~\eqref{eq:traj_velocity_limits}, acceleration~\eqref{eq:traj_acceleration_limits}, and torque limits~\eqref{eq:traj_torque_limits} are met, and for which all goals are reached in the intended order~\eqref{eq:traj_goal_order}.
We denote the set of all trajectories fulfilling these properties by:
\begin{subequations}\label{eq:trajectory_main_number}
    \begin{alignat}{2}
        \chi_T(R) = \{ &\qv(t): [0, t_{max}] \to \mathcal{Q} \mid \tag{\ref{eq:trajectory_main_number}} &&\\
                            &\mathcal{O} \cap \mathcal{A}(R, \qv(t)) = \emptyset \label{eq:traj_collision_free} && \\
                            &\land \dot{\qv}(t) \in \left[\, \dot{\ql}, \dot{\qu} \,\right] \label{eq:traj_velocity_limits} && \\
                            &\land \ddot{\qv}(t) \in \left[\, \ddot{\ql}, \ddot{\qu} \,\right] \label{eq:traj_acceleration_limits} && \\
                            &\land |\boldsymbol{\tau}| \leq \boldsymbol{\tau}_{max} \label{eq:traj_torque_limits} && \\
                            &\land \hat{r}(R, \qv(t), G, \mathcal{T} \}\, . \label{eq:traj_goal_order}
    \end{alignat}
\end{subequations}

If $\chi_T(R) \neq \emptyset$, we assume there is a path planning algorithm that computes a trajectory $\qv(t)$ for task $T$ with the robot $R$.
Under this assumption, we can define our objective function
\begin{align}
    C_T(R) = w_s C_s(R) + w_p C_{p, T}(R, \qv(t) ) \, , \label{eq:cost_function}
\end{align}
which is composed of a weighted sum of robot setup costs $C_s$ (such as acquisition cost or module availability) and process cost $C_{p, T}$ (such as cycle time or energy consumption) arising during operation.
Consequently, the robot defined by the optimal composition is given by
\begin{align}
    R_T^* &= \argmin_R(C_T(R)),~\text{s.t.}~\qv(t) \in \chi_T \, . \label{eq:optimization_objective}
\end{align}

\section{Method}
In a \ac{ga}, multiple solution candidates (\textit{chromosomes}) form the population, which is altered via the genetic operators in every generation.
To encode a manipulator as chromosome $c$ of fixed length $n_c$, we index the set of available modules $\mathcal{M}$ and write $c = ( m_1, \dots, m_{n_c} ), m_i \in \{0, 1, \dots, |\mathcal{M}|\}$ to encode the sequence of assembled module identifiers.
A gene encodes an empty slot when set to zero~\mbox{($m_i = 0$)} and a regular module from $\mathcal{M}$ otherwise.
In a slight abuse of notation, we will use a gene $m_i$ and the encoded module $M_i$ interchangeably in the remainder of this paper.
A series of genes $(u, 0, v)$ encodes a partial configuration of modules $(M_{u}, M_v)$.
This flexibility allows us to encode any manipulator with $n_M$ modules in a chromosome of consistent length $n_c \geq n_M$.
As a result, we neither predetermine the number of modules in a solution candidate nor constrain the search space to alternate joint and link modules, allowing us to explore unconventional compositions.

\subsection{Genetic Operators}
Within every generation, we compute the fitness value for every chromosome.
Based on the fitness, we perform steady state \textit{selection}: The $p$ solution candidates (population) with the highest fitness among all individuals are chosen to be the parents for the next generation.
We then replace any chromosome not selected by a new one created by a single-point \textit{crossover} between two randomly selected parents.
We denote the set of all modules with the same connectors by $\mathcal{V}_m$ and, finally, select a replacement candidate uniformly from $\mathcal{V}_m$ for every gene in the current population with a \textit{mutation} probability of $p_m$.
If the mutated chromosome $m_i$ is neither a base nor an end effector \mbox{($i \notin \{1, n_c\}$)}, the empty module is added to $\mathcal{V}_m$ if a connection between the adjacent modules $(m_{i-1}, m_{i+1})$ is possible.
Due to the inherent validity check of connections during population generation, we can incorporate constraints on module composition, such as those given by connector properties inherently, and before computing the fitness.
Therefore, even complex module libraries, e.g., modules of different sizes, can be optimized using our \ac{ga}.

\subsection{Fitness Function}
As the fitness function $f$ determines the quality of a solution candidate, we expect the relation \mbox{$C_T(R_i) < C_T(R_j) \Leftrightarrow f(R_i) > f(R_j)$} between fitness and task cost to hold.
However, due to the computationally expensive evaluation of the cost function in~\eqref{eq:cost_function} that requires attempting to compute a solution trajectory, it is infeasible to use it directly as a fitness function.
Given that our selection process relies on the order of fitness scores rather than their exact numerical values, we implement a lexicographic fitness function~\cite{Coello2000}:
We select a sequence of fitness objectives $f(R) = \langle f_1(R), \dots, f_n(R) \rangle$, ranked descending by importance and computational simplicity, and define the ordering
\begin{align}
    f(R_a) > f(R_b) \Leftrightarrow& \exists k \in \{1, \dots, n\}: f_k(R_a) > f_k(R_b) \nonumber \\
                & \land \forall i < k: f_i(R_a) = f_i(R_b)\, , \\
    f(R_a) = f(R_b) \Leftrightarrow& \forall i \in \{1, \dots, n\}: f_i(R_a) = f_i(R_b)\, .
\end{align}
As an illustrative example, let us consider two robots, $R_1$ and $R_2$, and evaluate them on two objectives.
The first objective, $f_1(R)$, is a binary classifier determining whether a trajectory $\qv(t) \in \chi_T(R)$ exists for each robot.
The second objective, \mbox{$f_2(R) = -n_J(R)$}, is defined as the negative of the number of joints for each robot.
Applying a lexicographic ordering, $R_1$ is preferred over $R_2$ if a trajectory exists for $R_1$ but not for $R_2$ or if a trajectory exists for both robots but $R_1$ has fewer joints than $R_2$.
Specifically, we introduce the following objectives for a robot $R$:
\begin{itemize}
    \item[1)] Trivial reachability: A goal can only be reached if a robot arm is sufficiently long. We introduce the maximum Euclidean distance between the distal and the proximal connector of module $M_i$ as $d_{max}(M_i)$ and define
        \begin{align} \label{eq:objective_f1}
            f_1(R) = \begin{cases}
                1 \text{, if } \sum_R d_{max}(M_i) \geq \max_{g \in G}{\vecabs\boldsymbol{p}_g\vecabs_2} \\ 
                0 \text{, otherwise.}
            \end{cases}
        \end{align}
    \item[2)] Reachable number of goals: Objective
        \begin{align} \label{eq:objective_f2}
            f_2(R) = \sum_{g \in G} \left( \begin{cases}
                1 \text{, if } \exists \qv \in \mathcal{Q}: r(R, \qv, g, \mathcal{T}) \\
                0 \text{, otherwise}
            \end{cases}\right)
        \end{align}
        returns the number of goal positions that can be reached by the robot $R$. The existence of a valid joint configuration $\qv$ is determined using a numeric inverse kinematics algorithm.
    \item[3)] Reachable number of goals, considering collisions: Objective
        \begin{align} \label{eq:objective_f3}
            f_3(R) = \sum_{g \in G} \left( \begin{cases}
                1 \text{, if } \exists \qv \in \mathcal{Q}: &r(R, \qv, g, \mathcal{T}) \\
                    &\land \mathcal{O} \cap \mathcal{A}(R, \qv) = \emptyset \\
                0 \text{, otherwise}
            \end{cases}\right)
        \end{align}
    returns the number of goal positions that can be reached while there are no collisions between robot and environment.
    \item[4)] Cost objective: The final fitness objective measures robot setup complexity and task performance and can be chosen freely. Following~\eqref{eq:cost_function},~\eqref{eq:optimization_objective}, we define
        \begin{align} \label{eq:objective_f4}
            f_4(R) = \begin{cases}
                -\infty, &\text{if } \chi_T(R) = \emptyset \\
                -C_T(R), &\text{otherwise.}
            \end{cases}
        \end{align}
\end{itemize}
We incrementally compute the lexicographic fitness value by subsequently evaluation these criteria and stopping if a composition does not achieve the maximum fitness value \mbox{$f_i(R) \neq f_{i, max}$} for an intermediate objective $i \leq 3$, where $f_{1, max} = 1$, \mbox{$f_{2, max} = f_{3, max} = |G|$}.
This can be interpreted as a hierarchical pruning procedure, backed by computational evidence about a robot's performance.
However, unlike prior work using hierarchical elimination to optimize \ac{modrob} compositions~\cite{Icer2017}, we retain information about partial solutions, such as the ability to reach the final goal in the task and use it for the selection process.
Moreover, the pruning relies exclusively on task performance, avoiding a human bias introduced by manually crafted rules and thereby preserving the explorative capabilities of \acp{ga}.
Finally, the introduction of a lexicographic fitness function enhances interpretability.
Rather than managing and fine-tuning weighted sums for numerous task criteria for a scalar fitness value, a human evaluator can easily conclude which constraints different solution candidates satisfy.

\section{Experiments}

\subsection{Robot Modules}
For our experiments, we used data from 29 different modules manufactured by RobCo\footnote{\url{https://www.robco.de/en}}.
The set consisted of four bases, 20 static link modules, four modules with a joint, and one end effector.
The static link modules differed in size and shape (L-shaped and I-shaped), and the bases differed in size and mounting orientation.
We divided the database into six large modules, including two bases, and 22 smaller modules, including the end effector that can only be attached to one of the large modules using a special connecting link module.
Without the constraints imposed by the connectors and the necessity of a base and end effector module, there would have been $\sum_{i=n}^{12} 29^n \approx 10^{17}$ distinct compositions with at most twelve modules that could be built from this module set.
Considering the different module sizes and types, the resulting size of the workspace was still beyond $10^{12}$ (one trillion) possible compositions for a chromosome length of twelve and, to the best of our knowledge, exceeds those in prior work, such as 15552 in~\cite{Icer2017}, 32768 in~\cite{Althoff2019}, and around $10^6$ in~\cite{Whitman2020} significantly.

\subsection{Task Definition}
\begin{figure*}
    \centering
    \subfloat[Synthetic I, $d =4$]{\includegraphics[width=0.245\linewidth,trim={650px 210px 650px 50px},clip]{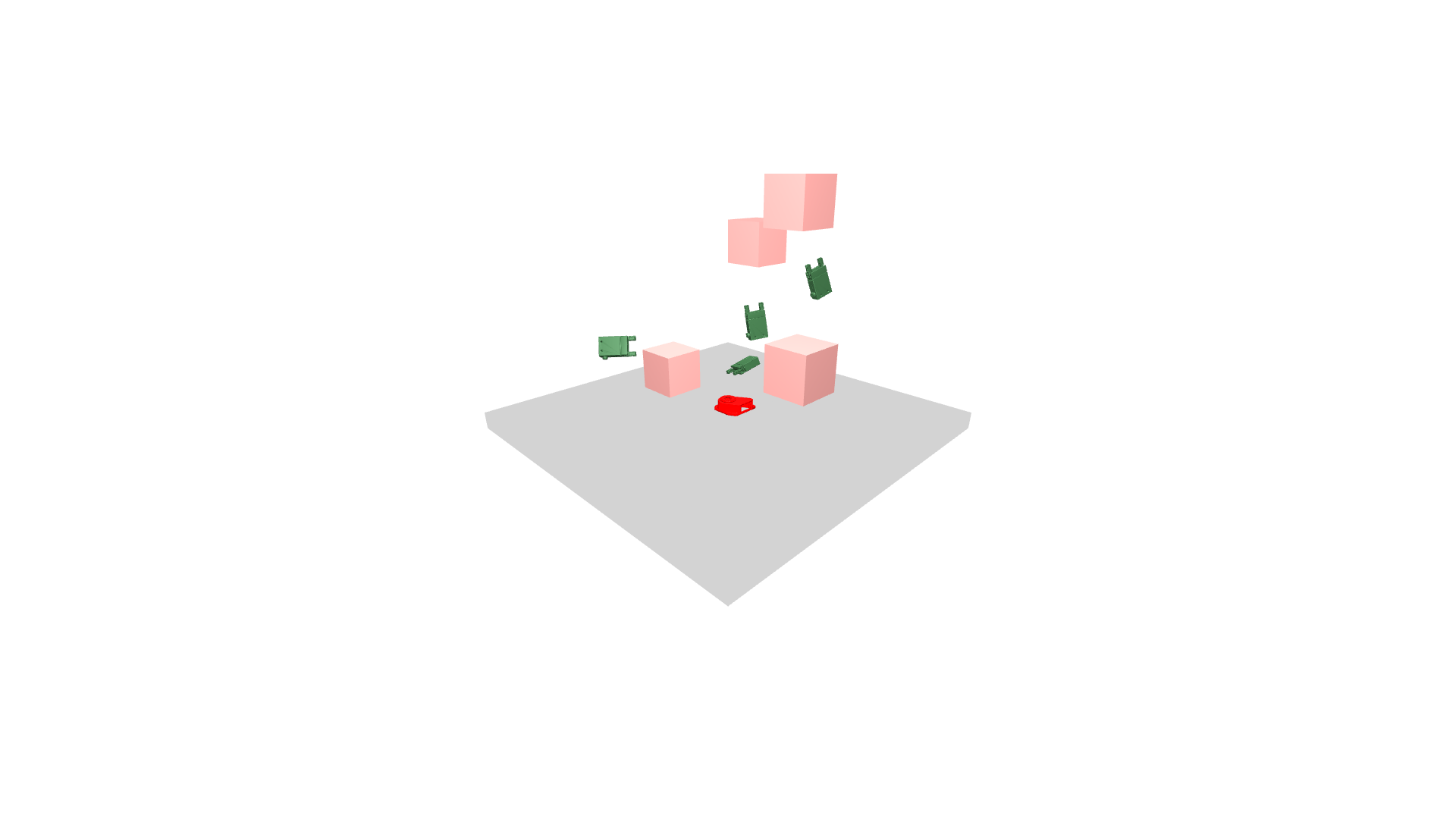}}
    \hfill
    \subfloat[Synthetic II, $d=3$]{\includegraphics[width=0.245\linewidth,trim={650px 210px 650px 50px},clip]{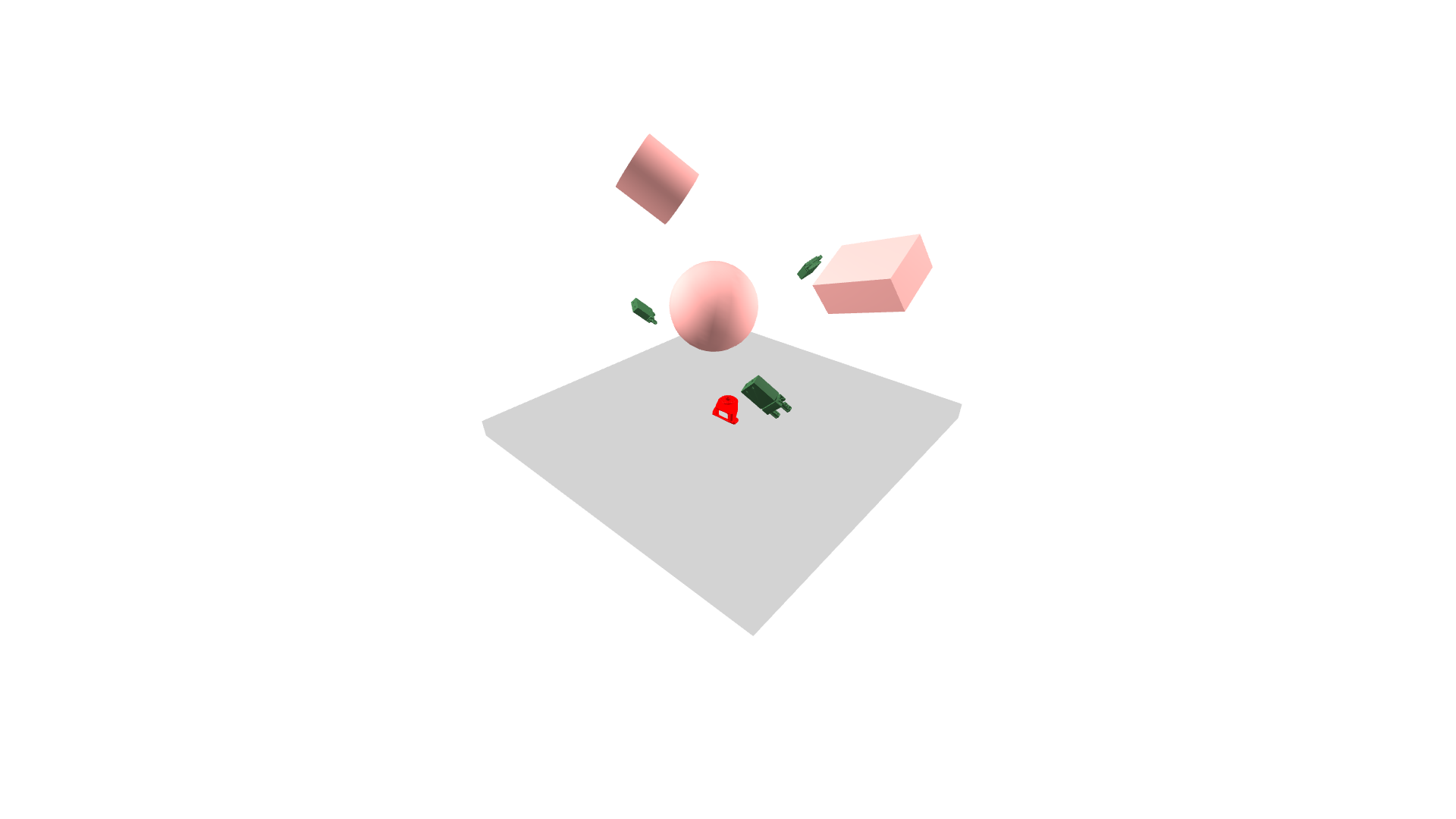}}
    \hfill
    \subfloat[Manufacturing I]{\includegraphics[width=0.245\linewidth,trim={450px 0px 450px 0px},clip]{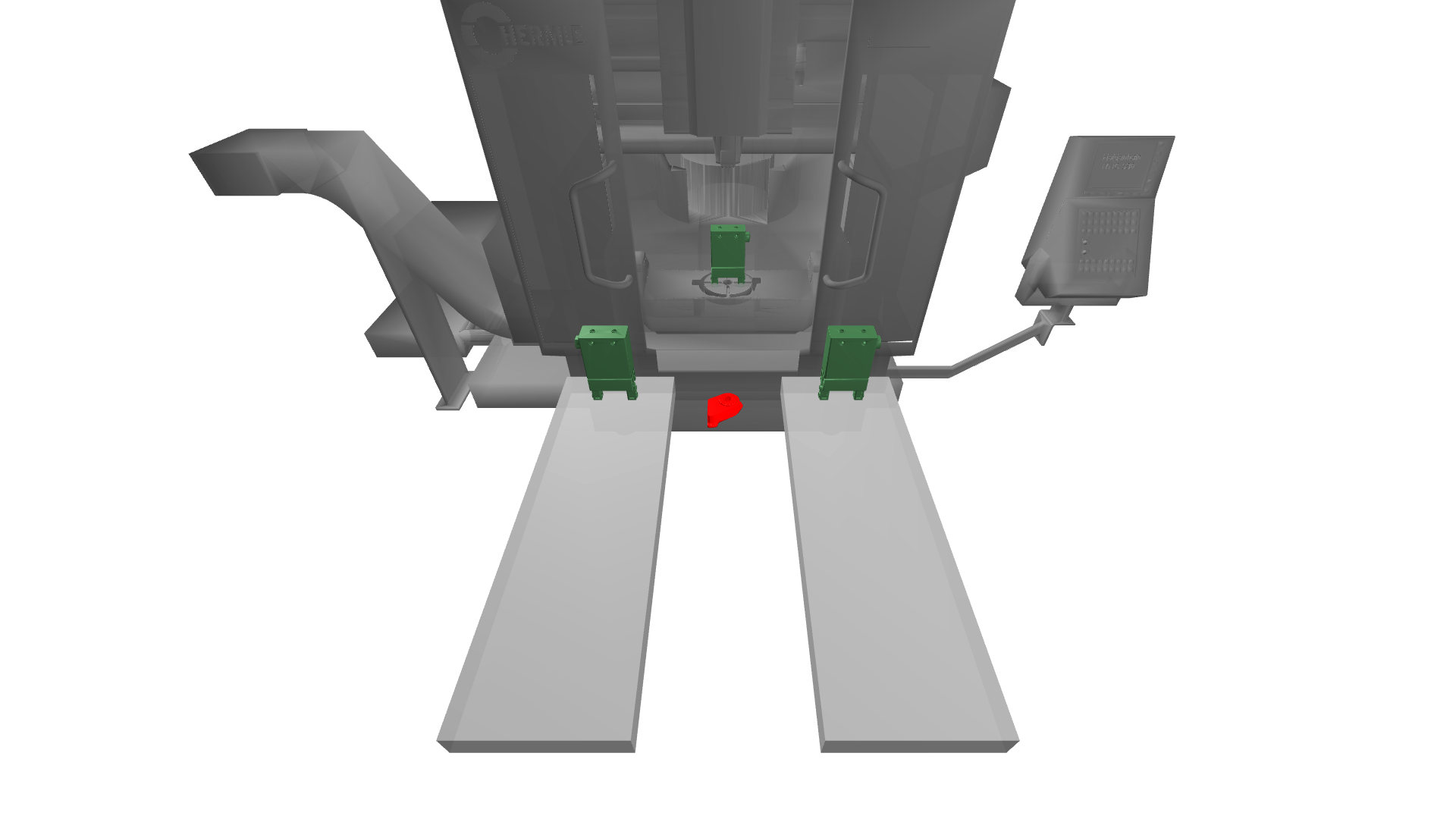}}
    \hfill
    \subfloat[Manufacturing II]{\includegraphics[width=0.245\linewidth,trim={520px 100px 520px 0px},clip]{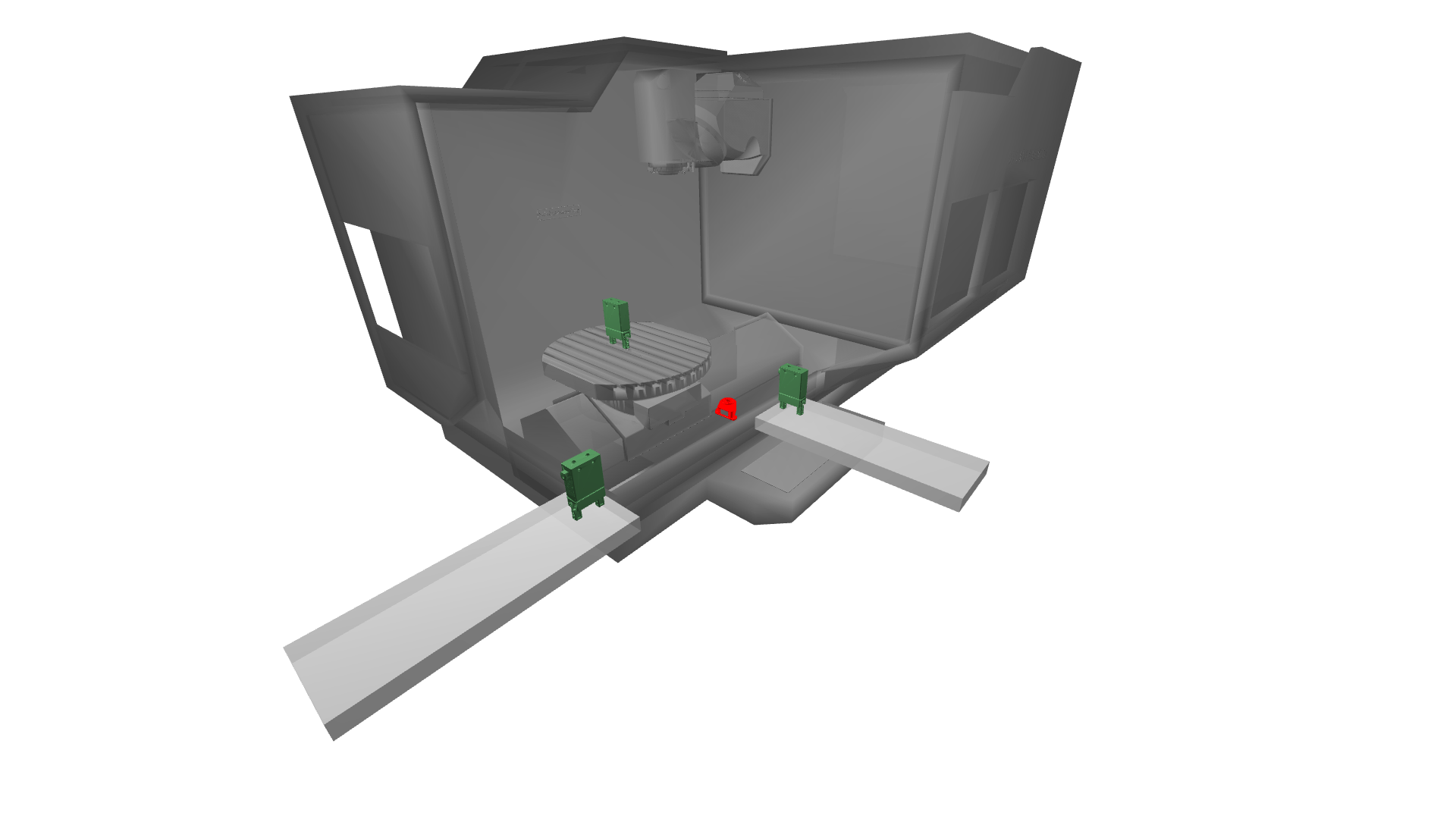}}
    \caption{We evaluated our algorithm on two synthetic and two manufacturing settings: Synthetic obstacles are shown in red, milling machines and conveyor belts are shown in grey. Every goal consists of a desired position and orientation, indicated by an end effector. The robot base placement is indicated by a red base module.}
    \label{fig:task_visualizations}
\end{figure*}
We evaluated our approach on two types of tasks: Randomly generated synthetic tasks and manually curated industrial manufacturing tasks.
We assessed the algorithm's performance for each type in two different settings and three difficulty levels each.
Fig.~\ref{fig:task_visualizations} shows an example for every type of task.

For the first setting (Synthetic I), we implemented a scenario similar to the one proposed in~\cite{Whitman2020} by discretizing a space of $1.25 \meter^3$, centered in the point $[0, 0, 0.625]^T \meter$ in $5 \times 5 \times 5$ voxels of edge width $0.25\meter$.
For the different difficulty levels, we randomly sampled $d \in \{3, 4, 5\}$ box-shaped obstacles, filling one voxel each and $d$ goal positions centered in a voxel each.
In addition, we created non-discretized synthetic tasks (Synthetic II), where goal and obstacle position and orientation were sampled randomly in a half-ball of radius $1.2m$ and with a positive $\za$ coordinate.
In this setting, obstacles could be spheres, boxes, or cylinders, and their position and volume were determined randomly.
For all goals in the synthetic tasks, we set the orientation tolerance to half a degree around an arbitrary axis \mbox{($\boldsymbol{e} = [1, 1, 1]^T, \varphi = \frac{\pi}{360}$)}.
Finally, we applied a simple, non-exhaustive heuristic to discard tasks that are not solvable, e.g., because obstacles covered one of the goal positions or the base position.
In total, we defined 120 synthetic tasks, 20 for each setting with $d \in \{3, 4, 5\}$.

Furthermore, we evaluated our approach in two settings inspired by real-world machine tending problems (Manufacturing I \& II), where the goals represent a pick position for raw material, a position inside a milling machine, and a position to place the final workpiece on.
For each one, a difficulty level was determined by using one of three orientation tolerances, mimicking different workpiece geometries:
\begin{itemize}
    \item Sphere-like geometry $(t_o = t_{o, 1})$: The \ac{tcp} orientation can be chosen freely within an arbitrary rotation of 90°, so we set \mbox{$\boldsymbol{e}_1  = [1, 1, 1]^T, \varphi_1 = \frac{\pi}{2}$}.
    \item Partially symmetric geometry $(t_o = t_{o, 2})$: The \ac{tcp} orientation is free around its $\za$-axis but otherwise constrained, so we set \mbox{$\boldsymbol{e}_2 = [\frac{1}{360}, \frac{1}{360}, 1]^T, \varphi_2 = \pi$}.
    \item Arbitrary geometry $(t_o = t_{o, 3})$: The \ac{tcp} orientation is pre-determined and we allow a minor deviation only, so we set \mbox{$\boldsymbol{e}_3 = [1, 1, 1]^T, \varphi_3 = \frac{\pi}{360}$}.
\end{itemize}
For all tasks, we set a position tolerance of $t_p = 10^{-3}\meter$.

\subsection{Baseline} \label{sec:baseline}
We compared our approach to a two-level \ac{ga} based on hierarchical elimination, as proposed in~\cite{Icer2017}.
We eliminated unfit solution candidates that could not reach the first and last goal as indicated by~\eqref{eq:objective_f3} before evaluation.
Instead of utilizing the previously introduced lexicographic fitness function, we adopted~\cite[(3)-(10)]{Icer2017}.
Consequently, the baseline fitness function depended on weighting factors $k_i$, linear distance $L$, angular distance $A$, dexterity $D$, number of modules $n_M$, number of joints $n_J$, joint value differences $V$, and the percentage of reachable intermediate points $P$:
\begin{align}
    f_B = e^{- \left( k_1 R + k_2 L + k_3 A + k_4 D + k_5 n_M + k_6 n_J + k_7 V \right)} + k_8 P\, . \label{eq:baseline_fitness}
\end{align}
We waived the obstacle proximity criterion initially introduced in the referenced work, which is defined solely for spherical obstacles, and replaced the involved module criterion $I$ by the number of links $n_M$ and the number of joints $n_J$.
All other criteria remained as defined in the reference.
In the second optimization step, we selected the best 25 individuals for each scenario and computed trajectories using the same method as employed for our solutions.

\subsection{Setup}
All experiments were conducted using Timor-Python~\cite{Kuelz2023} on a desktop PC with an Intel i7-11700KF processor.
For trajectory generation, we followed the method from~\cite{Kunz2012} in combination with the RRT-Connect planner provided by the \acf{ompl}~\cite{OMPL2012} with a timeout of 3 seconds.
To enhance the efficiency of evaluating recurring module compositions, we optimized the process by caching the kinematic and dynamic models derived from module data for the most recent 1000 robots assessed.
However, due to the stochastic nature of the RRT-Connect planner, we calculated the fitness value for each composition, regardless of whether it had been evaluated before.

We defined the optimization criterion as a function of the number of robot joints $n_J$, robot modules $n_M$, and trajectory cycle time $t_{max}$ as:
\begin{align} \label{eq:task_cost_function}
    C_T = n_J + 0.2 n_M + t_{max}\, .
\end{align}    
The algorithms ran for 200 generations on a population size of 25, which was initialized using weighted random sampling, where each initial module had a probability of $p_j = 0.9$ to contain a joint.
For evaluating the performance on the manufacturing tasks, we performed optimization over ten random seeds, whereas due to run time constraints, we ran optimization once on all synthetic tasks.
If not stated otherwise, all results refer to the mean computed over the difficulty levels $d \in \{3, 4, 5\}$ for synthetic tasks and for tolerances $t_{o, 1}$, $t_{o, 2}$, and $t_{o, 3}$ for the manufacturing tasks with differing seeds.

Additionally, to analyze the the ability of both algorithms to adapt to different preferences, we evaluated them in the ``Manufacturing II'' setting with tolerance $t_{o, 3}$ and cost functions %
\mbox{$C_T(w_J) = w_J n_J + (5 - w_J) t_{max}$}, %
where we evaluated both algorithms' performance over ten random seeds for weights \mbox{$w_J \in \{0, 1, \dots, 5\}$}.

\subsection{Results}
\begin{table}
    \vspace{1.7mm}
    \centering
    \caption{
        Summary of achieved tasks, average number of joints, and average cycle time.
        }
    \label{tab:results}
    \begin{tabular}{lccccc}
\toprule
 Task Setting &  & Achieved & $C_T$ & $n_J$ & $t_{max}$ \\
\midrule
\multirow{3}{*}{Synthetic I} & $d = 3$ & 16 & 11.86 & 6.2 & 3.1s \\
 & $d = 4$ & 10 & 15.16 & 6.8 & 5.7s \\
 & $d = 5$ & 10 & 16.42 & 6.7 & 7.1s \\
\cmidrule{1-6}
\multirow{3}{*}{Synthetic II} & $d = 3$ & 20 & 12.27 & 6.4 & 3.3s \\
 & $d = 4$ & 20 & 13.80 & 6.7 & 4.6s \\
 & $d = 5$ & 20 & 15.33 & 6.8 & 5.9s \\
\cmidrule{1-6}
\multicolumn{2}{l}{Synthetic  (average over all)}  & \textbf{16.0} & \textbf{14.14} & \textbf{6.6} & \textbf{5.0s} \\
\cmidrule{1-6}
Baseline &  & 15.7 & 19.61 & 8.0 & 14.9s \\
\cmidrule{1-6} \\[-12pt]
\cmidrule{1-6}
\multirow{3}{*}{Manufacturing I} & $t_{o, 1}$ & 10 & 12.81 & 5.3 & 5.0s \\
 & $t_{o, 2}$ & 10 & 14.33 & 6.7 & 5.0s \\
 & $t_{o, 3}$ & 10 & 15.05 & 6.6 & 5.8s \\
\cmidrule{1-6}
\multirow{3}{*}{Manufacturing II} & $t_{o, 1}$ & 10 & 10.15 & 4.5 & 3.4s \\
 & $t_{o, 2}$ & 10 & 12.48 & 6.4 & 3.6s \\
 & $t_{o, 3}$ & 10 & 12.27 & 6.0 & 3.8s \\
\cmidrule{1-6}
\multicolumn{2}{l}{Manufacturing  (average over all)}  & \textbf{10.0} & \textbf{12.85} & \textbf{5.9} & \textbf{4.5s} \\
\cmidrule{1-6}
Baseline &  & 10.0 & 18.17 & 7.8 & 12.9s \\
\bottomrule \\[-8pt]
\multicolumn{6}{l}{\scriptsize $C_T$: Cost as defined in eq.~\eqref{eq:cost_function}, $n_J$: Number of joints, $t_{max}$: Cycle time}
\end{tabular}

\end{table}
The evaluation took $960ms$ per individual on average, resulting in 1.2 hours for one optimization over 200 generations.
This is a significant speedup compared to determining the cost function for every individual and running into the path planner's 3s time limit for every unfit solution candidate.
Table~\ref{tab:results} reports the number of tasks achieved in different settings and the average lowest cost for all tasks with a valid solution.
In addition, we report the mean number of joints $n_J$ and cycle time $t_{max}$ for all solutions with minimal cost, as well as the aggregated mean over all synthetic and manufacturing tasks for our approach and the baseline according to section~\ref{sec:baseline}.

Our algorithm found a valid solution for 80\% of the synthetic tasks.
Upon manual inspection, we discovered that some remaining tasks are inherently unsolvable due to the proximity of goals and obstacles.
For the hand-curated industrial tasks, our optimization resulted in valid solutions for every trial.
On average, our algorithm yielded solutions with costs about 30\% lower than those generated by our baseline method.
For the synthetic tasks, our approach reduced the average number of joints by $1.4$ and decreased cycle times by 66\%.
In manufacturing tasks, we observed an average reduction of 1.9 joints and a decrease in cycle times by 65\%.
Notably, the robots' complexity changed with the task difficulty. 
Two different robots for the ``Manufacturing II'' setting with $t_o = t_{o, 1}$ and $t_o = t_{o, 3}$ are shown in Fig.~\ref{fig:trajectory_sol}.

\begin{figure}
    \centering
    \includegraphics[width=\linewidth]{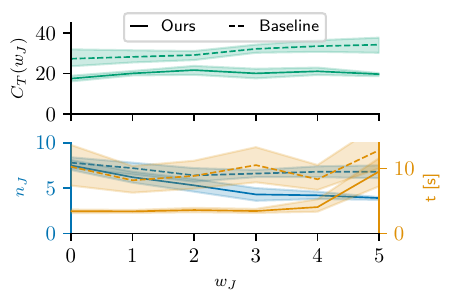}
    \caption{
    Our approach outperformed the baseline regardless of the weighting factor $w_J$. Especially for our approach, the tradeoff between $n_J$ and $t$ becomes visible.
    Shaded areas indicate 95\% confidence intervals computed using bootstrapping.
    }
    \label{fig:cost_weighting_impact}
\end{figure}
The effect of different cost functions on the optimization outcome is shown in Fig.~\ref{fig:cost_weighting_impact}.
For all weighting factors $w_J$, the proposed approach outperformed the baseline significantly.
The results of our approach show a tradeoff between the number of joints and the trajectory cycle time.
The baseline procedure found solutions with fewer joints if incentivized but failed to reduce the cycle time for lower $w_J$.
This result strongly indicates that including trajectory cycle time in the optimization process, as described in~\eqref{eq:objective_f4}, provides a significant advantage over the two-level approach deployed for the baseline.

\section{Conclusion}
This paper presents a genetic algorithm with a lexicographic fitness function for the task-based optimization of modular manipulators.
Compared to existing approaches, we impose no prior assumptions on the structure or the number of modules in an optimal solution.
By designing a computationally efficient fitness evaluation, our approach can handle complex scenarios and search spaces exceeding those presented in the related literature by magnitudes.
Our experimental validation shows that our \ac{ga} can find solutions adapted to the complexity of a task.
Moreover, the unconventional but high-performant module compositions resulting from optimizing \ac{modrob} for manufacturing tasks with broad orientation tolerances showcase that use-case-tailored \acp{modrob} do not necessarily follow human intuition.
Our approach applies to all kinds of serially connected \acp{modrob}, regardless of module complexity.
Additional task constraints can easily be integrated by extending the intermediate objectives evaluated during fitness computation.
Lastly, our method can be used for any task without the necessity of simplification, e.g., via discretization.

\section*{Acknowledgment}
The authors gratefully acknowledge financial support by the Horizon 2020 EU Framework Project CONCERT under grant 101016007.

\bibliographystyle{IEEEtran}
\bibliography{bibliography_files/bibliography_clean}

\end{document}